\documentclass[10pt,twocolumn,letterpaper]{article}

\usepackage{cvpr}              

\usepackage[dvipsnames]{xcolor}

\definecolor{cvprblue}{rgb}{0.21,0.49,0.74}
\usepackage[pagebackref,breaklinks,colorlinks,citecolor=cvprblue]{hyperref}

\usepackage{arydshln}
\usepackage{tabularx}
\usepackage{multirow}
\usepackage{makecell}
\usepackage{bbding}
\usepackage{xcolor}
\usepackage{colortbl}
\definecolor{darkred}{RGB}{139, 0, 0}
\definecolor{darkgreen}{RGB}{0,100,0}
\definecolor{lightblue}{rgb}{0.68, 0.85, 0.9}
\definecolor{darkblue}{rgb}{0.0, 0.0, 0.55}

\usepackage{array}

\usepackage{graphicx}
\usepackage{subcaption}
\usepackage{tabularx,ragged2e}
\newcolumntype{C}{>{\Centering\arraybackslash}X} 

\def\paperID{16142} 
\def\confName{CVPR}
\def\confYear{2024}

\title{NAVERO: Unlocking Fine-Grained Semantics for Video-Language Compositionality}

\author{
Chaofan Tao$^{1}$ \quad
Gukyeong Kwon$^{2}$ \quad
Varad Gunjal$^{2}$ \quad
Hao Yang$^2$ \quad
{Zhaowei Cai}$^2$ \\
{Yonatan Dukler}$^2$ \quad
{Ashwin Swaminathan}$^2$ \quad
{R. Manmatha}$^{2}$ \quad
{Colin Jon Taylor}$^{2}$ \quad
{Stefano Soatto}$^{2}$
\\
$^1$The University of Hong Kong \quad
$^2$AWS AI Labs\\
}
\begin{document}
\maketitle
\begin{abstract}
	
	We study the capability of Video-Language (VidL) models in understanding compositions between objects, attributes, actions and their relations. Composition understanding becomes particularly challenging for video data since the compositional relations rapidly change over time in videos. We first build a benchmark named AARO to evaluate composition understanding related to actions on top of spatial concepts. The benchmark is constructed by generating negative texts with incorrect action descriptions for a given video and the model is expected to pair a positive text with its corresponding video. Furthermore, we propose a training method called NAVERO which utilizes video-text data augmented with negative texts to enhance composition understanding. We also develop a negative-augmented visual-language matching loss which is used explicitly to benefit from the generated negative text. We compare NAVERO with other state-of-the-art methods in terms of compositional understanding as well as video-text retrieval performance. NAVERO achieves significant improvement over other methods for both video-language and image-language composition understanding, while maintaining strong performance on traditional text-video retrieval tasks.

\end{abstract}
\section{Introduction}
\label{sec:intro}


\begin{figure}[t]
	\centering
	\includegraphics[width=0.95\linewidth]{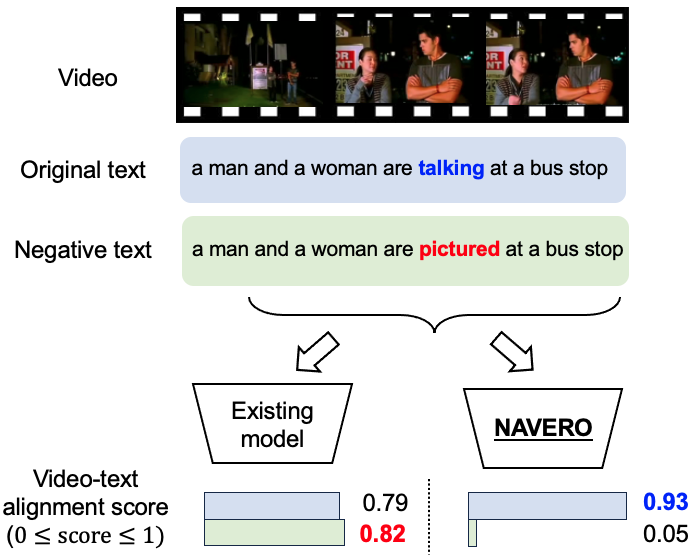}
	\caption{Video-Language compositional reasoning challenges the models to capture the subtle semantics in the captions. Given the paired video, the existing video-language models are confused in distinguishing the true caption from the corrupted captions, while NAVERO makes a clear distinction.}
	\label{fig:intro}
\end{figure}

Recent vision-language models have achieved remarkable performance in aligning visual data with corresponding text for tasks such as image classification and text-image retrieval~\citep{radford2021learning, li2022align, li2022blip}. Despite their success, it was also found that most of the existing image-text models are limited in understanding the composition between attributes, objects, and their relations~\citep{doveh2023dense, yuksekgonul2022and}. (e.g. an image with `man wearing \texttt{white} shoe ` can be easily misaligned with the text `man wearing \texttt{beige} shoe'). This limitation becomes more obvious with video data since the compositions presented in the scene vary over time. (e.g. a video with the scenes of `a man and a woman are \texttt{talking} at a bus stop' can be misaligned with the text `a man and a woman are \texttt{pictured} at a bus stop'). Figure~\ref{fig:intro} illustrates an actual failure case of a state-of-the-art method video-text model~\citep{cheng2023vindlu} in understanding the composition between an action and an object.  By making slight modifications to the original text, the model struggles to differentiate changes in the context of the paired video. Although there have been several attempts to address composition understanding issues in image-text models, the problem in the video domain has been largely overlooked.

One of the first required steps to overcome the failure on compositional reasoning is in building an evaluation benchmark to assess composition understanding capabilities of current models. Ideally, the benchmark should contain diverse negative texts which are plausible yet incorrect in describing a given video. However, annotating comprehensive and high quality negative texts for each video is expensive and time-consuming. Hence, inspired by existing works in the image domain~\citep{yuksekgonul2022and, doveh2023dense}, we utilize a mixture of rule-based and large language model (LLM)-based approaches to generate negative texts. We create a new benchmark named \textit{AARO} that contains corrupted negative captions, considering four compositional categories: `\textit{Action}', `\textit{Attribute}', `\textit{Relation}', and `\textit{Object}'  appearing in videos, to evaluate VidL models' compositional understanding on these semantics. Considering that videos contain rich temporal information unlike images, we construct an augmented word list that contains frequently appearing actions,\textit{ e.g.} `run', `walk'. 



We also propose a VidL training method, Negative-Augmented Video-language Enhancement for ReasOning (NAVERO), which improves composition understanding capability of models from both data and training objective aspects. For the data, we generate diverse hard-negative texts offline prior to the training process and use them as text augmentation for training. In generating hard negative texts, we conduct three main ablation studies to find the most effective generation approach. First, we explore different types of negative text generators such as rule-based, LLM-based, or mixture of both. Second, we evaluate the use of negative text generators in a multi-round fashion. Lastly, we analyze the importance of negative augmentation for actions to learn compositions along the temporal dimension. On the loss function side, we extend the negative-augmented image-text contrastive loss \citep{yuksekgonul2022and} to the domain of video-text, and introduce a new objective loss function, negative-augmented visual-text matching loss, that incorporates negative text in the visual-text matching loss \citep{li2021align}. The matching loss utilizes the generated negative text as hard-negatives and provides a direct guidance for compositional learning. To summarize, the main contributions of this paper are as follows:

\begin{itemize}
	\item  We build a comprehensive evaluation benchmark for assessing composition understanding capability of existing VidL models..
	
	\item We propose NAVERO, a method that utilize diverse generated negative texts and a negative-augmented visual-text matching objective to enhance composition understanding while maintaining performance on the text-video retrieval tasks.
	
	\item The video-text models trained by the proposed method show a good generalization to the image-text compositional tasks \citep{zhao2022vl}, achieving state-of-the-art (SOTA) performance with much less training data.
\end{itemize}

\section{Related Work}
\label{sec:related_work}

\subsection{Vision-Language Models}
Recent advancements in image-language models \citep{radford2021learning, jia2021scaling,tao2020dynamic,li2021align, yao2021filip, li2022blip, li2023blip, kwon2022masked, yu2022coca,wu2022nuwa,shi2023upop,shi2023crossget,wan2024electrocardiogram}, which learn from millions of web-crawled image-text pairs have led to strong performance improvements in various downstream applications including zero-shot image classification, image-text retrieval etc. Typically, a contrastive  objective function \citep{radford2021learning} is used to align the image features and the text features. Recent work \citep{li2022blip} indicates this design struggles with complex comprehension of language. Efforts to address these shortcomings have seen the development of models such as BLIP \citep{li2022blip} which proposes multi-objective training to enhance image-text learning for a better multimodal feature fusion. In parallel, large-scale video-language (VidL) models \citep{lei2021less, li2022align, bain2021frozen, wang2022omnivl, luo2022clip4clip, wang2023all, chen2022litevl, fu2021violet}  have demonstrated performance improvements in tasks such as text-video retrieval, video question answering, video captioning etc. VindLU \citep{cheng2023vindlu} empirically studies a variety of components proposed by prior works collectively and makes practical recommendations on how to build a highly performant VidL system.

\subsection{Visio-linguistic Compositionality}
For a comprehensive  understanding of visual scenes, it is crucial for vision-language models to gain a structured understanding of language and how it aligns with the visual content. This includes identifying entities and discerning how they relate to each other. This capability can unlock vision-language models in contributing to a broad range of applications such as action recognition \citep{ben2022bringing, wang2018videos}, scene graph \citep{xu2017scene, herzig2018mapping, krishna2018referring}, relational reasoning \citep{baradel2018object, battaglia2018relational} and human-object interactions \citep{kato2018compositional, xu2019learning}. To this end, the concept of compositional reasoning is traditinally proposed \citep{yuksekgonul2022and, zhao2022vl, thrush2022winoground, singh2023coarse, parcalabescu2021valse}  in learning structured information, \textit{e.g.} attribute, relation and object states appearing in the images. To learn the compositionality, prior works \citep{yuksekgonul2022and, doveh2023teaching} mainly utilize a variant of contrastive loss that considers the negative samples from the negative text augmentation or alternate images sampling. DAC \citep{doveh2023dense} also replaces the original text in the dataset by generating dense high quality text with auxiliary models. ICSVR \citep{madasu2023icsvr} evaluates how compositionality capabilities affect the performance of video retrieval. Orthogonally, we conduct an early study on improving video-language compositionality, with a comprehensive evaluation on video/image-text compositional reasoning and video retrieval tasks.



\section{Method}

\begin{figure*}[htp]
	\centering
	\includegraphics[width=0.83\linewidth]{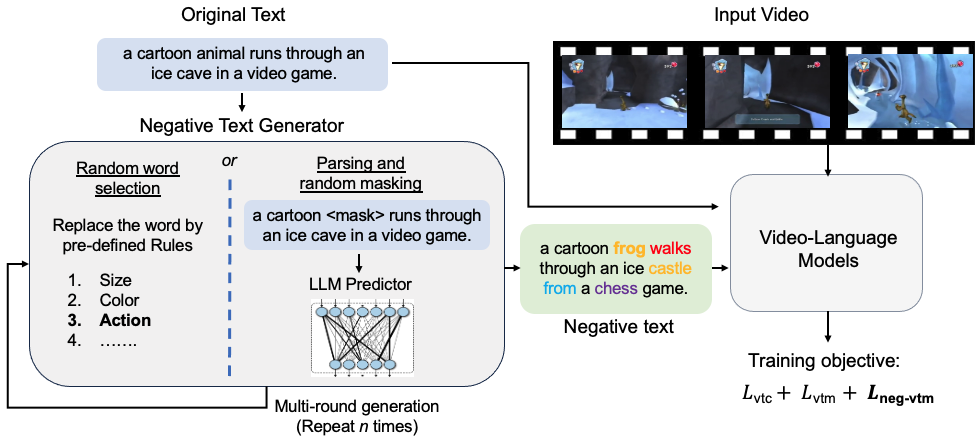}
	\caption{Illustration of the method NAVERO. We design a mixed-type multi-round  action-augmented negative text generator to create diverse negative texts. A new negative-augmented vision-text matching loss is proposed to boost the fine-grained compositionality learning.}
	\label{fig:framework}
	\vspace{-2mm}
\end{figure*}

We provide details on how we use negative text augmentation to build the AARO benchmark and generate diverse training data to boost compositional reasoning in Section \ref{sec:negative_text_aug}.  Given a source video-text dataset, we construct one negative text for each video-text pair. In Section \ref{sec:arch} we discuss the  architecture of the adopted VidL model. In Section \ref{sec:objectives}, we present the training objectives, including an extension of the negative loss used in image-text compositionality to the video-text domain, and a more effective negative-augmented vision-text matching loss. The illustration of the proposed NAVERO is presented in Figure \ref{fig:framework}.

\subsection{Negative Text Augmentation}
\label{sec:negative_text_aug}
\begin{table}
	\centering
	\scalebox{0.9}{
		\setlength{\tabcolsep}{1mm}{
			\begin{tabular}{c|c|c|c|c}
				\makecell{Negative\\Text Generator} & {Action} & {Attribute} & {Relation} & {Object}  \\
				\hline\hline
				Rule-based & action & \makecell{color, material,\\ state, size}  &relation & noun   \\
				\toprule
				LLM-based & VERB & ADJ & ADP  & NOUN   \\
			\end{tabular}
	}}
	\caption{The division to generate specific types (attribute, object, relation, action) of the negative texts. Take an example of generating attribute-type negative texts. In rule-basd approach, we replace the selected word with the words in the pre-defined words lists that contains color, material, state and size. In the LLM-based approach, we use parser to detect the adjective (ADJ), and then replace this word with the LLM predicted word.}
	\label{table:text_generators}
	\vspace{-2mm}
\end{table}

\subsubsection{Rule-based Text Augmentation}
Leveraging a set of pre-defined word lists, negative text augmentation can be effectively made by modifying the entities in the original text description.  To implement this, each targeted aspect, like color or material, has an associated word list. For each video, the original paired text is scanned for matches within this word list, and identified words are substituted with another word that is randomly chosen from the same category, thereby creating a negative pair. \citep{doveh2023teaching} provides a set of 
word and phrase lists which can be used as the rules for replacement.

\subsubsection{LLM-based Text Augmentation}
Another approach to create negative texts is based on Large Language Models (LLMs). Following previous work \citep{doveh2023teaching,doveh2023dense}, we adopt spaCy \citep{honnibal2017spacy} to parse a ground truth annotation into grammatical categories including nouns, verbs, adjectives (ADJ) and adpositions (ADP). Then, a particular  category and its word are chosen at random, masked, and substituted with a word predicted by the LLM's unmasking function.  The language model generally  provides plausible negative texts compared with the original text, thereby making it hard for the models to distinguish the substitution based on trivial characteristics.

\subsubsection{Diverse Negative Text Augmentation}
\label{sec:diverse_ntg}

\begin{figure}[htp]
	\vspace{-4mm}
	\centering
	\includegraphics[width=1\linewidth]{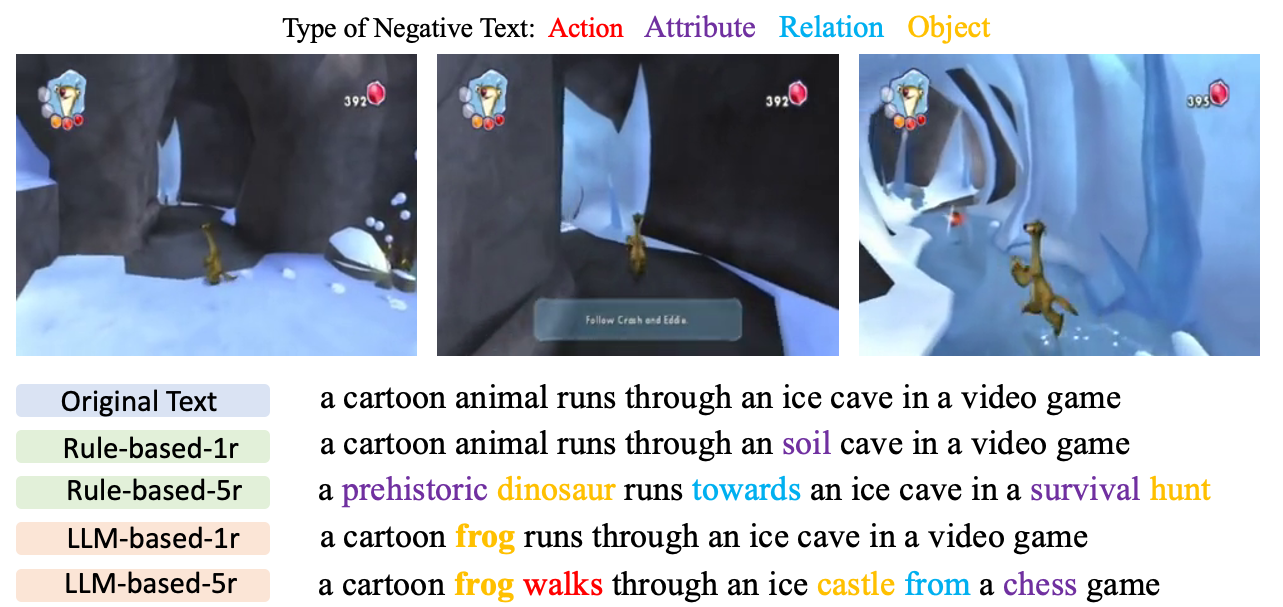}
	\caption{Example of a video-text pair where the text description is corrupted in a rule-based method and an LLM-based method. The suffixes `1r' and `5r' denote the number of rounds of calls to the negative text generator per sample. More rounds lead to more extensive set of variations from the original text.}
	\label{fig:example_augmention}
\end{figure}
Intuitively, the diversity of the negative text generation plays an important role in boosting the compositional understanding by improving generalization and reducing bias. To this end, we investigate 3 approaches for this: 1) Using multi-round negative text augmentation to replace additional words in a sentence, with the possibility of replacing all types (attribute, relation, object and action). After calling the text generator to replace one word in the previous text, we repeat the pipeline and pass the output text to the generator again until reaching the maximum number of rounds. 2) Using a mix of rule-based and LLM-based augmentation by randomly selecting one type of augmentation with equal probability. 3) By looking at frequently appearing actions in videos, we construct an action list of 273 common actions manually, and also add a relation list which describes inter-object spatial relations such as `in front of' and `behind'. For the detailed word lists in each category, please refer to the supplementary material.

The negative texts are generated offline before training. Examples of negative texts used for training are illustrated in Figure \ref{fig:example_augmention}. Section \ref{sec:ablation_text} presents how
these techniques of diverse augmentation affect the reasoning performance. To create the test data for isolated evaluation, as shown in Table \ref{table:text_generators}, we construct a series of datasets named AARO, containing 4 types of tasks, Action, Attribution, Relation and Object. In each task, only one type of negative text is assigned to each video-text pair.

\subsection{Model Architecture}
\label{sec:arch}
We use the ViT-B/16 \citep{dosovitskiy2020image} initialized by the weights of BEiT \citep{bao2021beit} pretrained on the ImageNet-21k as our video encoder. As for the temporal modelling for the video data, we adopt the divided space-time attention proposed by the TimeSformer \citep{bertasius2021space}, which computes the attention along the spatial tokens and temporal tokens respectively.  For the text encoder, we use a 12-layer BERT$_{base}$ \citep{devlin2018bert} with it's original initialization.  In addition, 3 cross-attention modules are inserted into the last 3 layers of the text encoder as suggested by VindLU \citep{cheng2023vindlu}, to enhance the video-text information fusion. A figure illustrating the model's architecture is illustrated in the supplementary material.

\subsection{Training Objectives}
\label{sec:objectives}

Each negative augmented video-text batch $B$ comprises of 3 quantities $\{ ( T^{neg}_i , T_i, V_i) \}$ where the $i$-th video $V_i$ is paired with one original text description $T_i$ and a negative description $T^{neg}_{i}$ that corrupts the words in $T_i$.  In this section, we provide the loss functions that we have experimented with in our work.

\paragraph{Vision-text contrastive loss} The text-video similarity score $S$ with the temperature $\sigma$ is computed as $S(T, V) = \exp( \frac{e^{T}_T e_V}{\sigma|e_T| |e_V|})$, where $e_T$, $e_V$ are the output embedding of the text and video, respectively.
$L_{vtc}$ is computed in standard form as
\begin{equation}
	L_{vtc}=\sum_i \log(\frac{S(T_i, V_i)}{\sum_j S(T_i, V_j)}) + \log(\frac{S(T_i, V_i)}{\sum_k S(T_k, V_i)}).
\end{equation}

\paragraph{Vision-text matching loss} We introduce
$L_{vtm}$, a vision-text matching loss term which provides binary predictions indicating whether a given video-text pair is aligned:
\begin{equation}
	L_{vtm} = \sum_i CE(y_{i}^{vtm}, p_{i}^{vtm}(T,V)), (T,V)\sim \{  (T_i, V_j) \},
\end{equation}
where $CE(\cdot)$ is the cross-entropy loss. For the $i$-th video-text pairs, $y_{i}^{vtm}$ is a 2D one-hot vector and $p_{i}^{vtm}$ denotes a two-class probability vector determining whether the input text-video pairs are matched. For each video in a mini-batch, typically the negative samples (non-matched text-video pairs $(T_i, V_j)$) are sampled from the same batch following the contrastive similarity distribution \citep{li2021align}, where texts that are more aligned to the video are viewed as the hard negatives and have a higher chance to be sampled.

\paragraph{Negative-augmented Vision-text contrastive loss} 
We extend the negative-augmented contrastive loss explored for image-text compositional reasoning \citep{doveh2023teaching} to video-text pairs. $L_{neg-vtc}$ is computed as -
\begin{equation}
	L_{neg-vtc} =-\sum_i \log(\frac{S(T_i, V_i)}{S(T_i, V_i)+S(T^{neg}_{i}, V_i)}).
\end{equation}

\paragraph{Negative-augmented Vision-text matching loss}  The goal of this loss function,  $L_{vtm-neg}$, is to train the model to differentiate between video-text pairs and video-negative text pairs. Rather than using random batch sampling for hard negative mining, we designate the video-negative text pairs $( T^{neg}_{i}, V_i)$ as hard-negatives. The intuition behind this is that negative texts, while seemingly plausible, provide incorrect descriptions of the video, making them ideal as hard-negatives samples. The $L_{neg-vtm}$ loss is formulated as follows -
\begin{equation}
	L_{neg-vtm} = \sum_i CE(y_{i}^{vtm}, p_{i}^{vtm}(T^{neg}_{i}, V_i)), 
\end{equation}
This loss is calculated atop the multimodal fusion module, thereby offering direct guidance for compositional learning. The fusion module is crucial as it models the interaction between video and text features, a key aspect in understanding compositional nuances.

We observe $L_{neg-vtm}$ provides better results than using $L_{neg-vtc}$ or the combination $L_{neg-vtc}+L_{neg-vtm}$. Therefore, we add the negative-augmented vision-text matching loss in NAVERO, addition to the vision-text contrastive loss and vision-text matching loss terms.


\section{Experiments}

\subsection{Training Procedure}
\label{sec:procedure}
As suggested by VindLU \citep{cheng2023vindlu}, we adopt CC3M \citep{sharma2018conceptual} and WebVid2M \citep{bain2021frozen} as our pre-training datasets.  Vision-text contrastive loss, vision-text matching loss and masked language modeling loss \citep{devlin2018bert} are used during the pre-training. During the fine-tuning stage of NAVERO, we adopt a mix of rule-based and LLM-based  generation to produce one negative text for each video-text pair. By default, each negative text is created in 5 rounds, iteratively modifying different aspect at each step. The hyper-parameters for the pre-training and fine-tuning are available in the supplementary material.

\subsection{Datasets}
\label{sec:datasets}

As explained in Section \ref{sec:diverse_ntg}, we use diverse negative text augmentation for the standard training split, and type-specific negative text generation for the testing split to evaluate compositional reasoning on each type in isolation. By using MSRVTT \citep{xu2016msr}, DiDeMo \citep{anne2017localizing} and ActivityNet as the source video-text datasets\citep{heilbron2015activitynet}, we construct a series of datasets with a prefix name `AARO' to evaluate the understanding of the action, attribute, relation and objects in each model. To study whether NAVERO can also boost image-text compositionality, we use a large-scale image-text compositional understanding benchmark VL-Checklist \citep{zhao2022vl}  and also construct the dataset AARO-COCO with image-text pairs sourced from the COCO captioning dataset \citep{lin2014microsoft}.  Further details for each dataset are available in the supplementary material.

\subsection{Evaluation Metric}
Following the image-text compositional reasoning work in VL-checklist \citep{zhao2022vl}, we use accuracy as the evaluation metric. The output of Video-Text Matching (VTM) head is adopted to compute the score \citep{zhao2022vl}. Given a video, the VidL model's output scores, are derived between the original text and the negative text. If the score associated with the original text description surpasses that of the negative text, this is viewed as a correct prediction. Accuracy (acc) is then determined based on a subsequent equation:
\begin{equation}
	acc = \frac{\sum_{i} f(T_i,T_i^{neg})}{N},
\end{equation}
where $N$ is the number of test samples, and $f(T_i,T_i^{neg})=1$ if $p(T_i|V_i) > p(T_i^{neg}|V_i)$. In addition, we notice the previously adopted metric does not guarantee the video-original text pairs are predicted positively and video-negative text pairs are predicted negatively. Therefore, we also consider a harder metric by adding the aforementioned constraint:
\begin{equation}
	\hat{acc} = \frac{\sum_{i} \mathbb{I} [p(T_i|V_i)> 0.5] }{2N}+\frac{\sum_{i} \mathbb{I} [p(T_i^{neg}|V_i) < 0.5] }{2N},
\end{equation}
where $\mathbb{I} [True]=1$ and  $\mathbb{I} [False]=0$.

\subsection{Compared Methods}
Since few prior works explore video-text compositional reasoning, we evaluate on 3 kinds of video-text models. 1) The first one (Vanilla) adopts $L_{vtc}$ and $L_{vtm}$ to directly fine-tune on the original downstream dataset without using any negative augmentation; 2) The second variant (NegVidL) adopts the technique of negative-augmented contrastive loss used in image-text compositionality  \citep{doveh2023teaching} on top of the Vanilla model; 3) The last model is fully trained with the proposed NAVERO that adopts proposed mixed-type multi-round action-augmented negative text augmentation, and the negative-augmented matching loss $L_{neg-vtm}$.


For image-text compositional reasoning, in order to study the generalization of our video-text models, we compare our approach with representative image-text models \textit{e.g.} CLIP 
\citep{radford2021learning}, BLIP2 \citep{li2023blip}, previous SOTA work on compositional reasoning \textit{e.g.} NegCLIP \citep{yuksekgonul2022and}, SVLC \citep{doveh2023teaching}, DAC \citep{doveh2023dense},and lastly our fine-tuned video-text and image-text models applied with NAVERO.


\begin{table*}[htp]
	\center
	\scalebox{0.95}{
		\setlength{\tabcolsep}{3.5mm}{
			\begin{tabular}{c|c|cccc|c}
				\textbf{Dataset} & Models & \textbf{Action}& \textbf{Attribute} & \textbf{Relation} & \textbf{Object}  & \textbf{Avg} \\
				\hline\hline
				\multirow{3}{*}{\centering AARO-MSRVTT} & Vanilla&  67.63/57.31 & 71.01/56.91 & 56.64/52.02 & 77.60/63.50  & 68.22/57.43 \\	
				&NegVidL & 68.53/57.92  & 71.01/57.58 & 57.77/54.17 & 78.00/63.75  & 69.39/58.36 \\
				&NAVERO &\textbf{79.58}/\textbf{60.49}& \textbf{82.98}/\textbf{65.43} & \textbf{84.70}/\textbf{67.70} & \textbf{83.90}/\textbf{67.85}  & \textbf{82.79}/\textbf{65.37} \\       
				\toprule    
				
				\multirow{3}{*}{\centering AARO-DiDeMo}  &Vanilla& 79.90/59.14 & 79.81/60.86 & 80.51/61.07 & 79.74/59.88  & 79.99/60.24 \\
				
				&NegVidL     & 87.88/80.20 & 87.32/79.99 &  87.76/81.28 & 88.12/80.69 &87.77/80.54  \\
				&NAVERO &\textbf{99.49}/\textbf{91.36}& \textbf{99.18}/\textbf{91.31} &\textbf{99.18}/\textbf{91.33} & \textbf{99.50}/\textbf{91.57} & \textbf{99.34}/\textbf{91.39} \\  
				
				\toprule    
				\multirow{3}{*}{\centering AARO-ActivityNet}& Vanilla& 
				87.63/62.59&88.64/64.54 &88.17/62.97 &88.29/64.32&88.18/63.61
				\\
				&NegVidL & 94.67/87.93& 95.12/89.04&94.87/88.83 & 94.87/88.55&94.88/88.59\\
				& NAVERO& \textbf{99.88/95.70}&\textbf{99.95/95.84}&\textbf{99.88/95.6}8& \textbf{99.90/95.64}&\textbf{99.90/95.72} \\    
				\bottomrule
			\end{tabular}
		}
	}
	\caption{Experimental results to examine the understanding of Action, Attribute, Relation and Object in the video-text compositionality on the datasets AARO-MSRVTT, AARO-DiDeMo and AARO-ActivityNet. The results are reported in the format of $acc$ / $\hat{acc}$}.
	\label{tbl:model_results}
	\vspace{-3mm}
\end{table*}

\subsection{Results on AARO datasets} 
\label{sec:aaro_main}

In Table \ref{tbl:model_results} we observe, for all datasets, the vanilla fine-tuned models that do not harness any negative augmentation achieve the lowest performance.  When comparing with NegVidL which adopts $\textit{L}_{neg-vtc}$, NAVERO  consistently shows higher average score across all datasets, suggesting that the proposed diverse negative augmentation and the negative-augmented matching loss offer a direct and significant improvement over the  negative-augmented contrastive loss. 

In  addition,  across all datasets, `Attribute' and `Object' categories generally showcase higher scores compared to the `Relation' and `Action' categories. This trend hints that the models find it relatively easier to capture attribute and object-based compositional details, whereas it is harder for the models to understand the changing inter-object relations and the dynamic actions appearing in the video. The tasks on AARO-DiDeMo and AARO-ActivityNet are less challenging than that on AARO-MSRVTT. We speculate the AARO-MSRVTT dataset has a lot of samples where one video matches multiple text descriptions, making it more difficult to detect the difference between original and negative text.

\section{Discussions}
\label{sec:discussions}

\subsection{Ablation on Negative Text Augmentation}
\label{sec:ablation_text}

\begin{table*}[htp]
	\centering
	\scalebox{0.95}{
		\setlength{\tabcolsep}{2mm}{
			\begin{tabular}{l|c|c|cccc|c}
				\hline
				\makecell{\textbf{Negative}\\\textbf{Text Generator}}
				& \makecell{\textbf{Number of}\\ \textbf{Rounds}} & 
				\makecell{\textbf{Augmented}\\ \textbf{Actions}} 
				& \textbf{Action} & \textbf{Attribute} & \textbf{Relation} & \textbf{Object} & \textbf{Avg} \\ 
				\hline\hline
				Mixed &  5 & $\checkmark$ &\textbf{79.58}/\textbf{60.49}& {82.98}/{65.43} & {84.70}/{67.70} & {83.90}/{67.85}  & \textbf{82.79}/\textbf{65.37}  \\ 
				\toprule
				Rule-based  & 5 & $\checkmark$& 76.34/59.77& \textbf{83.51}/\textbf{67.29}& 81.04/67.00 & 80.30/66.55& 80.30/65.15 \\ 
				LLM-based  & 5 & $\checkmark$  & 78.91/60.60 & 77.66/60.24 & 76.11/57.84 & 84.00/68.00& 79.17/61.67 \\ 
				Mixed  & 1 & $\checkmark$  & 78.71/57.69&	{83.10}/{65.40}&	\textbf{85.14}/\textbf{70.25}&	\textbf{84.05}/\textbf{68.05}&	{82.75}/{65.35} \\ 
				
				Mixed  & 5 & \scalebox{0.65}{\XSolid}&77.12/60.66	&83.24/64.76	& 83.82/67.00&	83.00/67.35& 81.80/64.94\\ 
				\bottomrule
			\end{tabular}
		}
	}
	\caption{Ablation study on the types of negative text augmentation. Results are reported on AARO-MSRVTT dataset.}
	\label{tbl:ablation_on_neg}
	\vspace{-3mm}
\end{table*}

\begin{table*}[htp]
	\center
	\scalebox{0.9}{
		\setlength{\tabcolsep}{2.5mm}{
			\begin{tabular}{cccc|cccc|c}
				$\textit{L}_{vtc}$& $\textit{L}_{vtm}$ &$\textit{L}_{neg-vtc}$&$\textit{L}_{neg-vtm}$ & \textbf{Action}& \textbf{Attribute} & \textbf{Relation} & \textbf{Object}  & \textbf{Avg} \\
				\hline\hline
				$\checkmark$ & $\checkmark$ &  & $\checkmark$ &\textbf{79.58}/\textbf{60.49}& {82.98}/\textbf{65.43} & \textbf{84.70}/\textbf{67.70} & \textbf{83.90}/\textbf{67.85}  & \textbf{82.79}/\textbf{65.37} \\      
				\toprule    
				$\checkmark$ & $\checkmark$ &   $\checkmark$ & $\checkmark$& 78.46/60.49 & \textbf{84.31}/65.16 & 84.45/67.64& \textbf{83.90}/67.35  & 82.78/65.16 \\
				$\checkmark$ & $\checkmark$ & $\checkmark$  & &
				69.42/58.31&70.48/58.11  &57.90/53.67&77.90/63.35 &68.93/58.36\\
				$\checkmark$ & $\checkmark$ &  & &  67.63/57.31 & 71.01/56.91 & 56.64/52.02 & 77.60/63.50  & 68.22/57.43 \\
				\bottomrule
			\end{tabular}
		}
	}
	\caption{Ablation study on the loss function during the fine-tuning. Results are reported on AARO-MSRVTT dataset.}
	\label{tbl:ablation_on_loss}
\end{table*}

As indicated in Table \ref{tbl:ablation_on_neg}, our study examines the impact of negative text augmentation during training. The findings show that using a mixed generator and combining both LLM-based and rule-based approaches for generating negative texts outperforms methods that rely on either generator exclusively. This outcome suggests that integrating multiple generators enhances compositional understanding. When the number of rounds is small such as using a single round as in previous work \citep{yuksekgonul2022and, doveh2023teaching}, the semantics of the original text and negative text is very similar, thereby making the negative text hard. This explains the performance of using 1-round negative text is competitive. We observe however, that using 5-round negative text can also boost the performance with diverse information in the negative text. Additionally, we noted a decline in performance when the action word list is not expanded, particularly in action-type evaluations. This finding emphasizes that augmenting actions is useful for effectively understanding action-related compositional information in videos. 

\vspace{-0.2cm}
\subsection{Ablation on the Loss Functions}
\label{sec:ablation_loss}
We present the ablation of the loss function during fine-tuning in Table \ref{tbl:ablation_on_loss}. Compared with the Vanilla model, both the negative-augmented vision-text contrastive loss and negative-augmented vision-text matching loss improve the compositionality. The negative-augmented vision-text matching loss has more significant influence on the results than the contrastive loss. Intuitively, a video-text model with good compositionality is expected to have in-depth understanding for the fine-grained semantics.  As discussed in \citep{li2021align}, the contrastive loss is computed on the visual features and text features learned in their own spaces, making it hard to understand the semantics from the multimodal interaction. Compared with negative-augmented contrastive loss, the  negative-augmented matching loss is added on top of the multimodal fusion module and based on a more grounded vision-language representation learning. Therefore, the proposed negative-augmented matching loss provides more powerful support on the compositional understanding than the negative-augmented contrastive loss.

\subsection{Generalization to Image-Text Data}

\begin{table*}[ht]
	\centering
	\scalebox{0.85}{
		\setlength{\tabcolsep}{4.2mm}{
			\begin{tabular}{l|cc|ccc|c}
				\hline
				\textbf{Model} & \makecell{\textbf{Vision-Language}\\ \textbf{Pre-Training Data}} & \makecell{\textbf{Negative Augmented}\\\textbf{Fine-tuning Data}} & \textbf{Attribute} & \textbf{Relation} & \textbf{Object} & \textbf{Avg} \\
				\hline
				\rowcolor{lightblue} \multicolumn{7}{c}{\color{black}Previous Image-Text Models} \\        
				
				CLIP & 400M & 0 & 67.60 & 63.05 & 81.58 & 70.74 \\
				BLIP2 & 129M & 0 & 80.12 & 70.72 & 84.14 & 78.33 \\
				NegCLIP & 400M & 567K (COCO) & 73.24 & 62.53 & 81.36 & 72.38 \\
				SVLC & 400M & 3M (CC3M) & 71.97 & 68.95 & 85.00 & 75.31 \\
				DAC & 400M & 3M (CC3M) & 77.27  & \textbf{86.41}& 87.30 & 83.66 \\
				\toprule
				\rowcolor{lightblue} \multicolumn{7}{c}{\color{black}Our Video-Text Models} \\       
				Pretrained & 5M & 0 & 55.90 & 59.95 & 47.95 & 54.60 \\
				NAVERO  & 5M & 180K(AARO-MSRVTT) & 77.03  & 83.91& 87.90 & 82.95 \\
				\toprule
				\rowcolor{lightblue} \multicolumn{7}{c}{\color{black}Our Image-Text Models} \\       
				NAVERO  & 5M & 567K(AARO-COCO) & \textbf{81.29}& {84.11}  & \textbf{90.88} & \textbf{85.42} \\
				\bottomrule
			\end{tabular}
	}}
	\caption{Comparison of the recent work in image-text compositional reasoning in accuracy $acc$, the pre-trained video-text model, and the fine-tuned video-text models and image-text models using the proposed NAVERO on the VL-Checklist dataset.}
	\vspace{-2mm}
	\label{table:performance_vlcheck}
\end{table*}

Table \ref{table:performance_vlcheck} presents the performance between previous image-text models and our models on the VL-Checklist dataset. Despite CLIP's extensive training on hundreds of millions of vision-language data points, its compositional understanding is relatively weak. BLIP2 outperforms CLIP possibly due to its multimodal fusion module which enhances the interaction between visual and textual features, thereby improving reasoning. NegCLIP and SVLC use the negative augmentation on the COCO and CC3M datasets respectively, based on the pre-trained CLIP. The performance gap between  NegCLIP, SVLC and CLIP demonstrate the positive effect of negative augmentation. DAC advances this further by employing an additional image captioner and an LLM to generate high-quality captions for each image.  

To adapt NAVERO to image data, we treat the image as the single-frame video. Our video-text models excel in image-text compositional reasoning, achieving good results with significantly less vision-language pre-training and negative augmented fine-tuning data.  While the pre-trained model initially show limitations in image-text compositionality, NAVERO enables the model to reach competitive performance on the VL-checklist with reduced data. It suggests the representations learned through the video-text compositionality have promising transferability on the image-text data. Moreover, by applying NAVERO on the COCO image-text dataset which has only 567K image-text pairs, our model achieves state-of-the-art performance, highlighting the effectiveness of NAVERO in the image-text domain.



				\subsection{Forgetting Problems when Training for Compositional Reasoning}
				We examine whether training with negative augmentation hurts the performance on traditional text-video retrieval task. This is important since we may encounter catastrophic forgetting when training vision-language models with new concepts \citep{doveh2023teaching, castro2018end, kemker2018measuring}. It is important to preserve the original abilities when adapting to new data. We study this issue by fine-tuning our Vanilla models on the standard text-video retrieval datasets, and also fine-tune our models based on NAVERO. To illustrate the competitiveness of our models, we also compare with previous SOTA video-text models, including ClipBert \citep{lei2021less}, ALPRO \citep{li2022align}, Frozen \citep{bain2021frozen}, OmniVL \citep{wang2022omnivl}, VIOLET \citep{fu2021violet}, All-in-one \citep{wang2023all}
				and CLIP4CLIP \citep{luo2022clip4clip}.
				
				Table \ref{tbl:model_on_retrieval} presents our findings. Our model consistently demonstrates superior performance across various datasets compared to previous methods. Crucially, the incorporation of negative augmentation does not lead to catastrophic forgetting. When we enhance compositional reasoning through fine-tuning with negative-text augmentation, our model's retrieval performance is on par, or even exceeds that of the original model. This result illustrates the effectiveness of our approach in simultaneously improving compositional reasoning and preserving the original model functionalities.

				\begin{table*}[hpt]
					\centering
					\scalebox{0.8}{
						\setlength{\tabcolsep}{3mm}{
							\begin{tabular}{c|cccl|cccl|cccl}
								\textbf{Model}  & \multicolumn{4}{c}{\textbf{MSRVTT}}
								& \multicolumn{4}{c}{\textbf{DiDeMo}} 
								& \multicolumn{4}{c}{\textbf{ActivityNet}}\\
								\hline\hline
								& R1 & R5 & R10 & Avg & R1 & R5 & R10 & Avg & R1 & R5 & R10 & Avg \\
								\toprule
								
								\rowcolor{lightblue} \multicolumn{13}{c}{\color{black}State-of-the-art Video-Text Models} \\
								
								ClipBert	&	22.0&	46.8	&59.9& 42.9&	20.4&	48.0	&60.8	&43.1&21.3&	49.0	&63.5&44.6\\
								ALPRO	&	33.9&	60.7	&73.2&55.9&	35.9	&67.5&	78.8&	60.7& -&	-&	-& \\
								Frozen &	31.0	&59.5	&70.5	&53.7&34.6&	65.0	&74.7	& 58.1&-&	-	&-& \\
								OmniVL	& 47.8	&74.2	&83.8&	68.6&52.4&	79.5	&85.4&72.4&	-	&-	&-&\\
								VIOLET	&	34.5	&63.0	&73.4&57.0	&32.6&	62.8	&74.7	&56.7&-	&-&	-&\\
								All-in-one	&	37.9	&68.1	&77.1	&61.2&32.7	&61.4&	73.5&55.9	&22.4&	53.7&	67.7&47.9\\
								CLIP4Clip	&	44.5	&71.4&	81.6&65.8&42.8	&68.5	&79.2&	63.5&40.5	&72.4	&83.4& 65.4\\
								
								\toprule
								\rowcolor{lightblue} \multicolumn{13}{c}{\color{black}Our Video-Text Models} \\        
								Vanilla & 42.8 & 69.2 & 79.3 & 63.8 & 53.4 & 81.1 & 88.9 & 74.5 & 51.1 & 78.4 & 87.8 & 72.4\\
								\hline
								NAVERO  & 43.1&	70.0&	79.5 & 64.2$_{\textcolor{darkred}{\uparrow 0.4}}$ & 54.2 & 80.5 & 87.6 & 74.1$_{\textcolor{darkgreen}{\downarrow 0.4}}$& 50.7 & 78.5 & 87.2 & 72.1$_{\textcolor{darkgreen}{\downarrow 0.5}}$ \\
								\bottomrule
							\end{tabular}
						}
					}
					\caption{Performance on text-video retrieval tasks. We compare with the State-of-the-art video-text models, and also study whether  the involvement of the proposed negative augmentation bring the problem of catastrophic forgetting that
						hurts the retrieval performance.}
					\vspace{-1mm}
					\label{tbl:model_on_retrieval}
				\end{table*}

				\begin{figure}[hpt]
					\centering
					\begin{subfigure}[b]{0.23\textwidth}
						\centering
						\includegraphics[width=\textwidth]{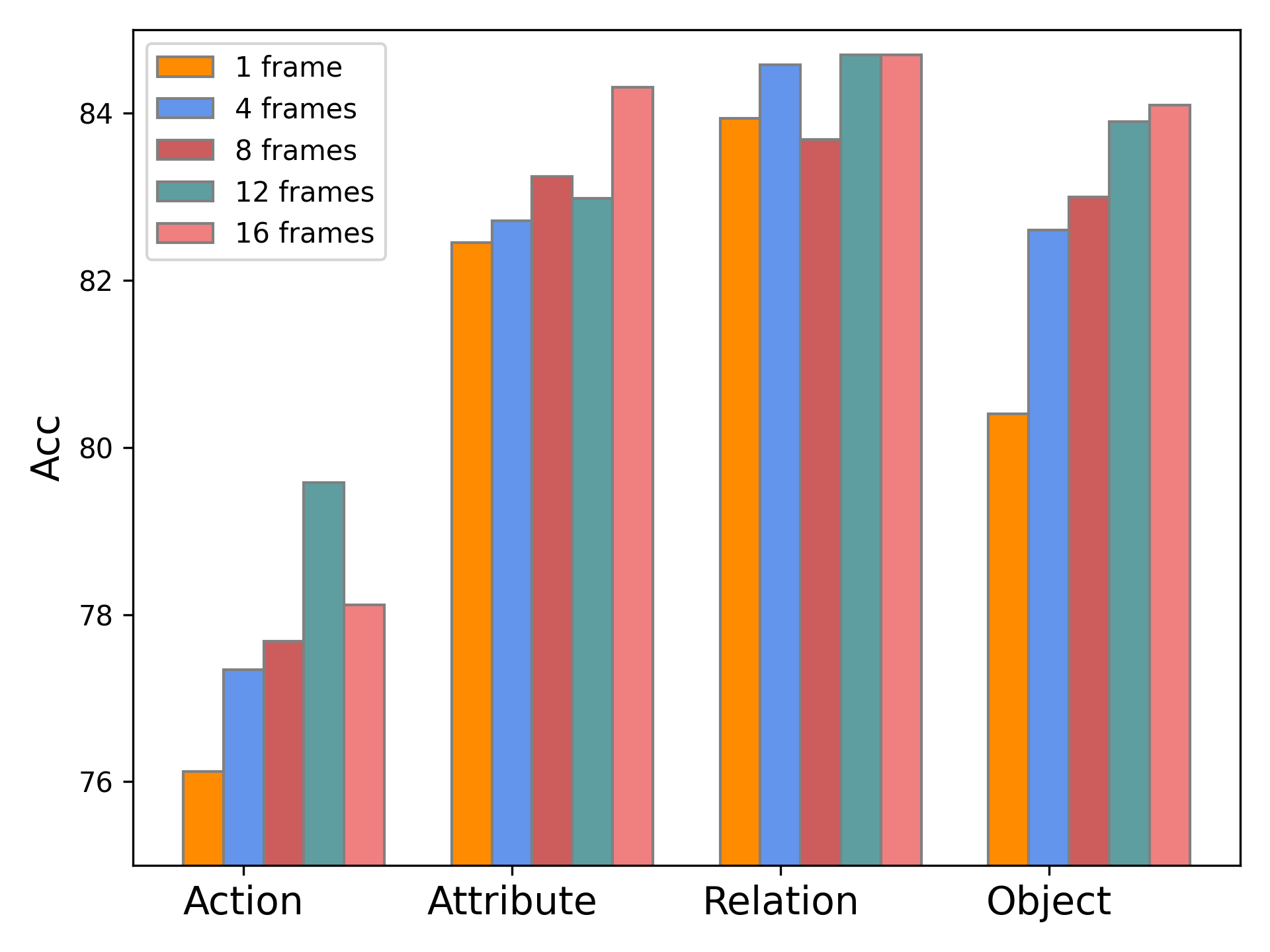}
						\caption{AARO-MSRVTT dataset}
						\label{fig:frame-sub1}
					\end{subfigure}
					\hfill
					\begin{subfigure}[b]{0.23\textwidth}
						\centering
						\includegraphics[width=\textwidth]{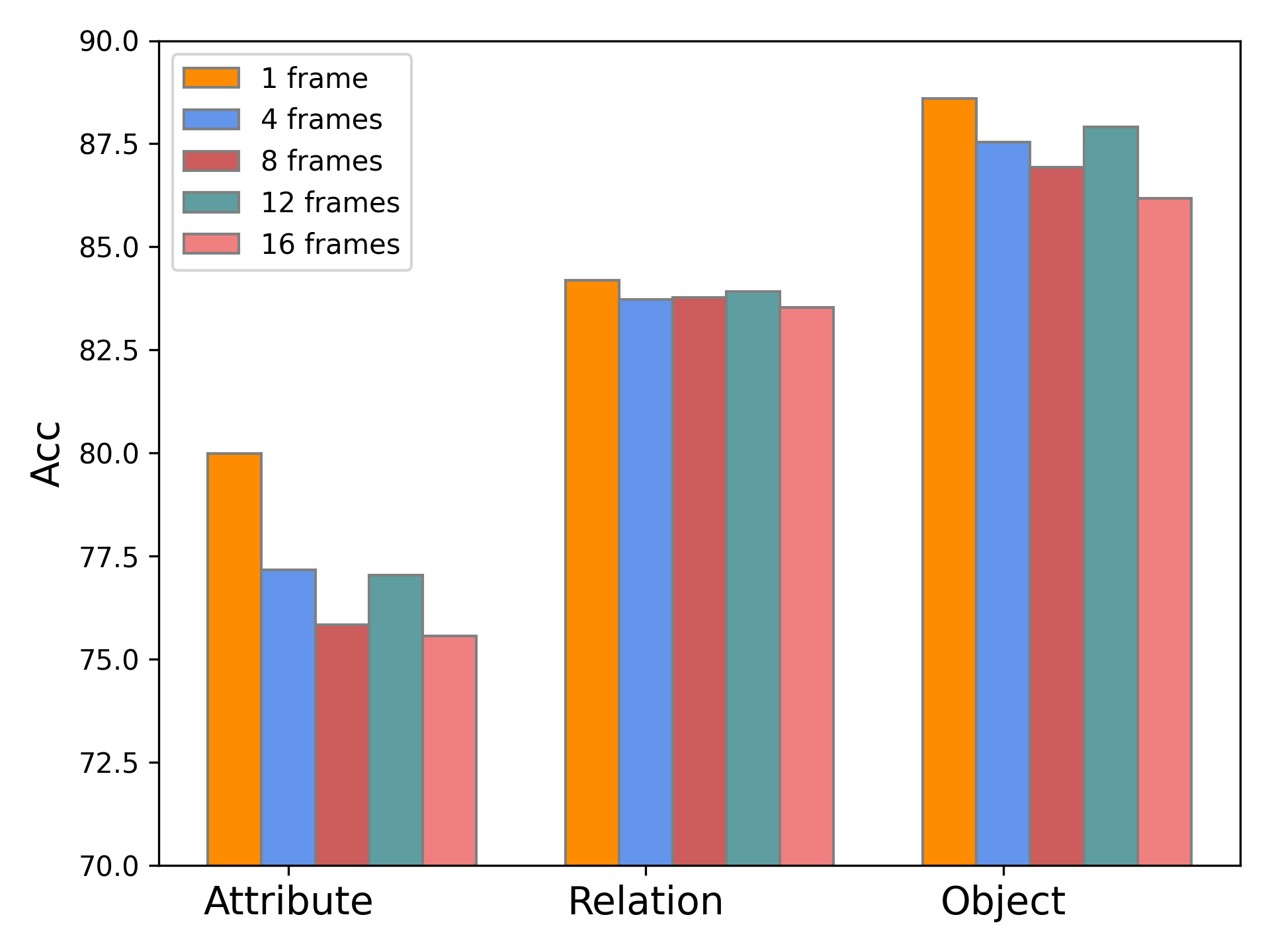}
						\caption{VL-Checklist dataset}
						\label{fig:frame-sub2}
					\end{subfigure}
					\caption{Bar plot  of accuracy $acc$ on each compositional type of evaluation on the video-text and image-text compositionality with the video-text model trained on different number of frames. Results are reported on AARO-MSRVTT and VL-Checklist dataset.} 
					\vspace{-1mm}
					\label{fig:ablation_frame}
				\end{figure}
				
					
						
				\begin{figure*}[htp]
					\centering
					\includegraphics[width=0.9\linewidth, height=7cm]{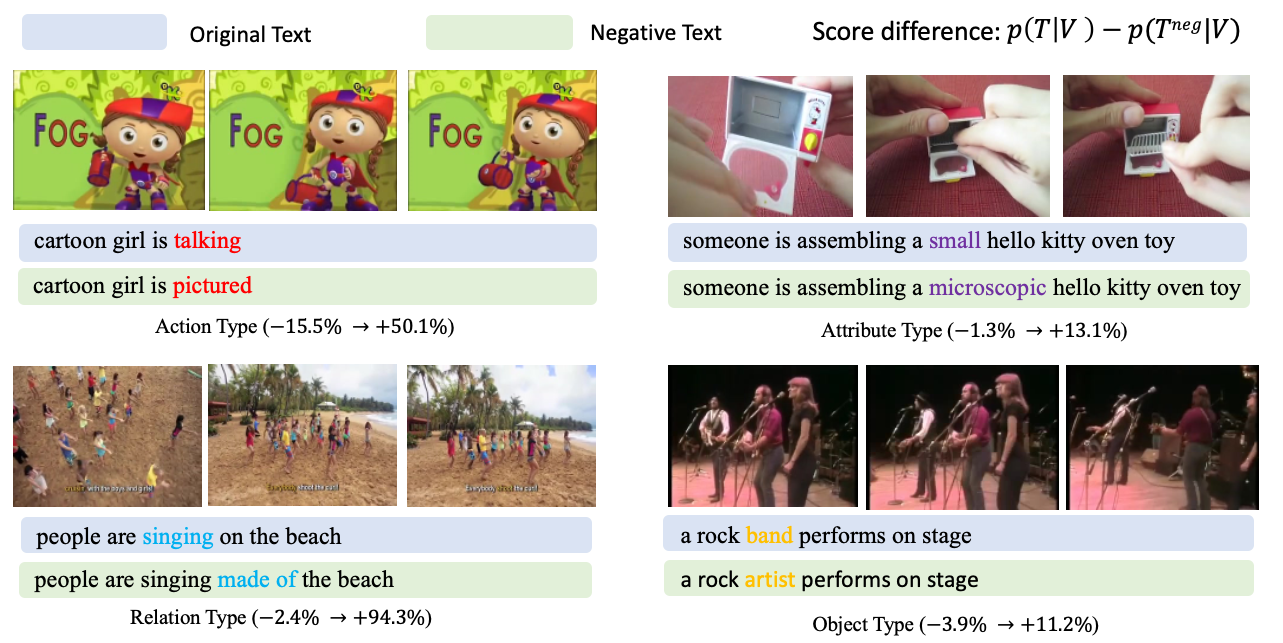}
					\caption{Case study for training on the Vanilla and NAVERO model on the AARO-MSRVTT. We provide test examples which are incorrectly predicted by the vanilla video-text model but are correctly predicted by the NAVERO. The score difference $(p(T|V) - p(T^{neg}|V))$ from Vanilla to NAVERO are reported, where $T$, $T^{neg}$ and $V$ denote the original text, negative text and the video, respectively.}
					\label{fig:case_study}
				\end{figure*}

				\subsection{Ablation on the Number of Frames}
				We study how the number of frames used per video during fine-tuning affect the compositional reasoning capability of the model. In Figure \ref{fig:ablation_frame}, we experimented with varying the number of frames per video (1, 4, 8, 12, and 16) while fine-tuning the pre-trained video-text model on the AARO-MSRVTT training split, we then evaluate the model on the AARO-MSRVTT and VL-Checklist datasets. For the video-text  compositional reasoning,  more frames indicate more dense temporal semantics, boosting the reasoning especially for the action composition category. Intriguingly, for image-text compositional reasoning, treating the video as a static image (1 frame) results in the best performance, despite the fact that all fine-tuned models are showing significant improvement over the pre-trained model in Table \ref{table:performance_vlcheck}. This indicates that while high-quality video-text datasets downstream can enhance the image-text compositionality generally,  a video-text model's image comprehension diminishes as it becomes increasingly focused on the temporal aspects of video data.
				\vspace{-0.6cm}
				\subsection{Case Study}
				\vspace{-0.2cm}
				We present a case study in Figure \ref{fig:case_study}. For each evaluation type, we showcase a sample that the Vanilla model misclassified, but is correctly identified by NAVERO. When the text  undergoes minor alterations, the Vanilla model struggles to discern differences. Even for the negative text which has an unrealistic description ``people are singing \texttt{made of} the beach'', the Vanilla model predicts a higher scores than that of the original text. In contrast, NAVERO effectively differentiates between the original and the negative text, demonstrating its effectiveness in compositional reasoning.

				
\section{Conclusion}\vspace{-0.15cm}
The introduced AARO benchmark and the NAVERO method contribute to enhanced compositional reasoning  in VidL models. AARO facilitates comprehensive assessments of VidL capabilities in understanding fine-grained semantics in videos, including actions, attributes, relations and object categories. NAVERO utilizes negative text augmentation and the negative-augmented vision-text matching loss to bolster compositional understanding. NAVERO outperforms the studied baselines on the AARO benchmark consistently for different video-text compositional datasets, and maintains good performance on text-video retrieval without catastrophic forgetting. In addition, NAVERO presents strong generalization on image-text compositional benchmarks, underlining its effectiveness and applicability to a broader range of vision-language tasks.

{
    \small
    \bibliographystyle{ieeenat_fullname}
    \bibliography{main}

\begin{thebibliography}{55}
\providecommand{\natexlab}[1]{#1}
\providecommand{\url}[1]{\texttt{#1}}
\expandafter\ifx\csname urlstyle\endcsname\relax
  \providecommand{\doi}[1]{doi: #1}\else
  \providecommand{\doi}{doi: \begingroup \urlstyle{rm}\Url}\fi

\bibitem[Bain et~al.(2021)Bain, Nagrani, Varol, and Zisserman]{bain2021frozen}
Max Bain, Arsha Nagrani, G{\"u}l Varol, and Andrew Zisserman.
\newblock Frozen in time: A joint video and image encoder for end-to-end
  retrieval.
\newblock In \emph{Proceedings of the IEEE/CVF International Conference on
  Computer Vision}, pages 1728--1738, 2021.

\bibitem[Bao et~al.(2021)Bao, Dong, Piao, and Wei]{bao2021beit}
Hangbo Bao, Li Dong, Songhao Piao, and Furu Wei.
\newblock Beit: Bert pre-training of image transformers.
\newblock \emph{arXiv preprint arXiv:2106.08254}, 2021.

\bibitem[Baradel et~al.(2018)Baradel, Neverova, Wolf, Mille, and
  Mori]{baradel2018object}
Fabien Baradel, Natalia Neverova, Christian Wolf, Julien Mille, and Greg Mori.
\newblock Object level visual reasoning in videos.
\newblock In \emph{Proceedings of the European Conference on Computer Vision
  (ECCV)}, pages 105--121, 2018.

\bibitem[Battaglia et~al.(2018)Battaglia, Hamrick, Bapst, Sanchez-Gonzalez,
  Zambaldi, Malinowski, Tacchetti, Raposo, Santoro, Faulkner,
  et~al.]{battaglia2018relational}
Peter~W Battaglia, Jessica~B Hamrick, Victor Bapst, Alvaro Sanchez-Gonzalez,
  Vinicius Zambaldi, Mateusz Malinowski, Andrea Tacchetti, David Raposo, Adam
  Santoro, Ryan Faulkner, et~al.
\newblock Relational inductive biases, deep learning, and graph networks.
\newblock \emph{arXiv preprint arXiv:1806.01261}, 2018.

\bibitem[Ben~Avraham et~al.(2022)Ben~Avraham, Herzig, Mangalam, Bar, Rohrbach,
  Karlinsky, Darrell, and Globerson]{ben2022bringing}
Elad Ben~Avraham, Roei Herzig, Karttikeya Mangalam, Amir Bar, Anna Rohrbach,
  Leonid Karlinsky, Trevor Darrell, and Amir Globerson.
\newblock Bringing image scene structure to video via frame-clip consistency of
  object tokens.
\newblock \emph{Advances in Neural Information Processing Systems},
  35:\penalty0 26839--26855, 2022.

\bibitem[Bertasius et~al.(2021)Bertasius, Wang, and
  Torresani]{bertasius2021space}
Gedas Bertasius, Heng Wang, and Lorenzo Torresani.
\newblock Is space-time attention all you need for video understanding?
\newblock In \emph{ICML}, page~4, 2021.

\bibitem[Castro et~al.(2018)Castro, Mar{\'\i}n-Jim{\'e}nez, Guil, Schmid, and
  Alahari]{castro2018end}
Francisco~M Castro, Manuel~J Mar{\'\i}n-Jim{\'e}nez, Nicol{\'a}s Guil, Cordelia
  Schmid, and Karteek Alahari.
\newblock End-to-end incremental learning.
\newblock In \emph{Proceedings of the European conference on computer vision
  (ECCV)}, pages 233--248, 2018.

\bibitem[Chen et~al.(2022)Chen, Tao, Hou, Shang, Jiang, and
  Liu]{chen2022litevl}
Dongsheng Chen, Chaofan Tao, Lu Hou, Lifeng Shang, Xin Jiang, and Qun Liu.
\newblock Litevl: Efficient video-language learning with enhanced
  spatial-temporal modeling.
\newblock \emph{arXiv preprint arXiv:2210.11929}, 2022.

\bibitem[Cheng et~al.(2023)Cheng, Wang, Lei, Crandall, Bansal, and
  Bertasius]{cheng2023vindlu}
Feng Cheng, Xizi Wang, Jie Lei, David Crandall, Mohit Bansal, and Gedas
  Bertasius.
\newblock Vindlu: A recipe for effective video-and-language pretraining.
\newblock In \emph{Proceedings of the IEEE/CVF Conference on Computer Vision
  and Pattern Recognition}, pages 10739--10750, 2023.

\bibitem[Devlin et~al.(2018)Devlin, Chang, Lee, and Toutanova]{devlin2018bert}
Jacob Devlin, Ming-Wei Chang, Kenton Lee, and Kristina Toutanova.
\newblock Bert: Pre-training of deep bidirectional transformers for language
  understanding.
\newblock \emph{arXiv preprint arXiv:1810.04805}, 2018.

\bibitem[Dosovitskiy et~al.(2020)Dosovitskiy, Beyer, Kolesnikov, Weissenborn,
  Zhai, Unterthiner, Dehghani, Minderer, Heigold, Gelly,
  et~al.]{dosovitskiy2020image}
Alexey Dosovitskiy, Lucas Beyer, Alexander Kolesnikov, Dirk Weissenborn,
  Xiaohua Zhai, Thomas Unterthiner, Mostafa Dehghani, Matthias Minderer, Georg
  Heigold, Sylvain Gelly, et~al.
\newblock An image is worth 16x16 words: Transformers for image recognition at
  scale.
\newblock \emph{arXiv preprint arXiv:2010.11929}, 2020.

\bibitem[Doveh et~al.(2023{\natexlab{a}})Doveh, Arbelle, Harary, Alfassy,
  Herzig, Kim, Giryes, Feris, Panda, Ullman, et~al.]{doveh2023dense}
Sivan Doveh, Assaf Arbelle, Sivan Harary, Amit Alfassy, Roei Herzig, Donghyun
  Kim, Raja Giryes, Rogerio Feris, Rameswar Panda, Shimon Ullman, et~al.
\newblock Dense and aligned captions (dac) promote compositional reasoning in
  vl models.
\newblock \emph{arXiv preprint arXiv:2305.19595}, 2023{\natexlab{a}}.

\bibitem[Doveh et~al.(2023{\natexlab{b}})Doveh, Arbelle, Harary, Schwartz,
  Herzig, Giryes, Feris, Panda, Ullman, and Karlinsky]{doveh2023teaching}
Sivan Doveh, Assaf Arbelle, Sivan Harary, Eli Schwartz, Roei Herzig, Raja
  Giryes, Rogerio Feris, Rameswar Panda, Shimon Ullman, and Leonid Karlinsky.
\newblock Teaching structured vision \& language concepts to vision \& language
  models.
\newblock In \emph{Proceedings of the IEEE/CVF Conference on Computer Vision
  and Pattern Recognition}, pages 2657--2668, 2023{\natexlab{b}}.

\bibitem[Fu et~al.(2021)Fu, Li, Gan, Lin, Wang, Wang, and Liu]{fu2021violet}
Tsu-Jui Fu, Linjie Li, Zhe Gan, Kevin Lin, William~Yang Wang, Lijuan Wang, and
  Zicheng Liu.
\newblock Violet: End-to-end video-language transformers with masked
  visual-token modeling.
\newblock \emph{arXiv preprint arXiv:2111.12681}, 2021.

\bibitem[Heilbron et~al.(2015)Heilbron, Escorcia, Ghanem, and
  Niebles]{heilbron2015activitynet}
Fabian~Caba Heilbron, Victor Escorcia, Bernard Ghanem, and Juan~Carlos Niebles.
\newblock Activitynet: A large-scale video benchmark for human activity
  understanding.
\newblock In \emph{Proceedings of the IEEE conference on Computer Vision and
  Pattern Recognition}, pages 961--970, 2015.

\bibitem[Hendricks et~al.(2017)Hendricks, Wang, Shechtman, Sivic, Darrell, and
  Russell]{anne2017localizing}
Lisa~Anne Hendricks, Oliver Wang, Eli Shechtman, Josef Sivic, Trevor Darrell,
  and Bryan Russell.
\newblock Localizing moments in video with natural language.
\newblock In \emph{Proceedings of the IEEE International Conference on Computer
  Vision}, pages 5803--5812, 2017.

\bibitem[Herzig et~al.(2018)Herzig, Raboh, Chechik, Berant, and
  Globerson]{herzig2018mapping}
Roei Herzig, Moshiko Raboh, Gal Chechik, Jonathan Berant, and Amir Globerson.
\newblock Mapping images to scene graphs with permutation-invariant structured
  prediction.
\newblock \emph{Advances in Neural Information Processing Systems}, 31, 2018.

\bibitem[Honnibal(2017)]{honnibal2017spacy}
M Honnibal.
\newblock Spacy 2: Natural language understanding with bloom embeddings,
  convolutional neural networks and incremental parsing, sentometrics research,
  2017.

\bibitem[Jia et~al.(2021)Jia, Yang, Xia, Chen, Parekh, Pham, Le, Sung, Li, and
  Duerig]{jia2021scaling}
Chao Jia, Yinfei Yang, Ye Xia, Yi-Ting Chen, Zarana Parekh, Hieu Pham, Quoc Le,
  Yun-Hsuan Sung, Zhen Li, and Tom Duerig.
\newblock Scaling up visual and vision-language representation learning with
  noisy text supervision.
\newblock In \emph{International conference on machine learning}, pages
  4904--4916. PMLR, 2021.

\bibitem[Kato et~al.(2018)Kato, Li, and Gupta]{kato2018compositional}
Keizo Kato, Yin Li, and Abhinav Gupta.
\newblock Compositional learning for human object interaction.
\newblock In \emph{Proceedings of the European Conference on Computer Vision
  (ECCV)}, pages 234--251, 2018.

\bibitem[Kemker et~al.(2018)Kemker, McClure, Abitino, Hayes, and
  Kanan]{kemker2018measuring}
Ronald Kemker, Marc McClure, Angelina Abitino, Tyler Hayes, and Christopher
  Kanan.
\newblock Measuring catastrophic forgetting in neural networks.
\newblock In \emph{Proceedings of the AAAI conference on artificial
  intelligence}, 2018.

\bibitem[Krishna et~al.(2017)Krishna, Zhu, Groth, Johnson, Hata, Kravitz, Chen,
  Kalantidis, Li, Shamma, Bernstein, and Fei-Fei]{krishna2017vg}
Ranjay Krishna, Yuke Zhu, Oliver Groth, Justin Johnson, Kenji Hata, Joshua
  Kravitz, Stephanie Chen, Yannis Kalantidis, Li-Jia Li, David~A Shamma,
  Michael Bernstein, and Li Fei-Fei.
\newblock Visual genome: Connecting language and vision using crowdsourced
  dense image annotations.
\newblock In \emph{International Journal of Computer Vision}, 2017.

\bibitem[Krishna et~al.(2018)Krishna, Chami, Bernstein, and
  Fei-Fei]{krishna2018referring}
Ranjay Krishna, Ines Chami, Michael Bernstein, and Li Fei-Fei.
\newblock Referring relationships.
\newblock In \emph{Proceedings of the IEEE conference on computer vision and
  pattern recognition}, pages 6867--6876, 2018.

\bibitem[Kwon et~al.(2022)Kwon, Cai, Ravichandran, Bas, Bhotika, and
  Soatto]{kwon2022masked}
Gukyeong Kwon, Zhaowei Cai, Avinash Ravichandran, Erhan Bas, Rahul Bhotika, and
  Stefano Soatto.
\newblock Masked vision and language modeling for multi-modal representation
  learning.
\newblock \emph{arXiv preprint arXiv:2208.02131}, 2022.

\bibitem[Lei et~al.(2021)Lei, Li, Zhou, Gan, Berg, Bansal, and
  Liu]{lei2021less}
Jie Lei, Linjie Li, Luowei Zhou, Zhe Gan, Tamara~L Berg, Mohit Bansal, and
  Jingjing Liu.
\newblock Less is more: Clipbert for video-and-language learning via sparse
  sampling.
\newblock In \emph{Proceedings of the IEEE/CVF conference on computer vision
  and pattern recognition}, pages 7331--7341, 2021.

\bibitem[Li et~al.(2022{\natexlab{a}})Li, Li, Li, Niebles, and
  Hoi]{li2022align}
Dongxu Li, Junnan Li, Hongdong Li, Juan~Carlos Niebles, and Steven~CH Hoi.
\newblock Align and prompt: Video-and-language pre-training with entity
  prompts.
\newblock In \emph{Proceedings of the IEEE/CVF Conference on Computer Vision
  and Pattern Recognition}, pages 4953--4963, 2022{\natexlab{a}}.

\bibitem[Li et~al.(2021)Li, Selvaraju, Gotmare, Joty, Xiong, and
  Hoi]{li2021align}
Junnan Li, Ramprasaath Selvaraju, Akhilesh Gotmare, Shafiq Joty, Caiming Xiong,
  and Steven Chu~Hong Hoi.
\newblock Align before fuse: Vision and language representation learning with
  momentum distillation.
\newblock \emph{Advances in neural information processing systems},
  34:\penalty0 9694--9705, 2021.

\bibitem[Li et~al.(2022{\natexlab{b}})Li, Li, Xiong, and Hoi]{li2022blip}
Junnan Li, Dongxu Li, Caiming Xiong, and Steven Hoi.
\newblock Blip: Bootstrapping language-image pre-training for unified
  vision-language understanding and generation.
\newblock In \emph{International Conference on Machine Learning}, pages
  12888--12900. PMLR, 2022{\natexlab{b}}.

\bibitem[Li et~al.(2023)Li, Li, Savarese, and Hoi]{li2023blip}
Junnan Li, Dongxu Li, Silvio Savarese, and Steven Hoi.
\newblock Blip-2: Bootstrapping language-image pre-training with frozen image
  encoders and large language models.
\newblock \emph{arXiv preprint arXiv:2301.12597}, 2023.

\bibitem[Li et~al.(2019)Li, Xu, Liu, Huang, Xu, Chen, Ma, Wang, Fang, and
  Lu]{li2019hake}
Yong-Lu Li, Liang Xu, Xinpeng Liu, Xijie Huang, Yue Xu, Mingyang Chen, Ze Ma,
  Shiyi Wang, Hao-Shu Fang, and Cewu Lu.
\newblock Hake: Human activity knowledge engine.
\newblock \emph{arXiv preprint arXiv:1904.06539}, 2019.

\bibitem[Lin et~al.(2014)Lin, Maire, Belongie, Hays, Perona, Ramanan,
  Doll{\'a}r, and Zitnick]{lin2014microsoft}
Tsung-Yi Lin, Michael Maire, Serge Belongie, James Hays, Pietro Perona, Deva
  Ramanan, Piotr Doll{\'a}r, and C~Lawrence Zitnick.
\newblock Microsoft coco: Common objects in context.
\newblock In \emph{Computer Vision--ECCV 2014: 13th European Conference,
  Zurich, Switzerland, September 6-12, 2014, Proceedings, Part V 13}, pages
  740--755. Springer, 2014.

\bibitem[Luo et~al.(2022)Luo, Ji, Zhong, Chen, Lei, Duan, and
  Li]{luo2022clip4clip}
Huaishao Luo, Lei Ji, Ming Zhong, Yang Chen, Wen Lei, Nan Duan, and Tianrui Li.
\newblock Clip4clip: An empirical study of clip for end to end video clip
  retrieval and captioning.
\newblock \emph{Neurocomputing}, 508:\penalty0 293--304, 2022.

\bibitem[Madasu and Lal(2023)]{madasu2023icsvr}
Avinash Madasu and Vasudev Lal.
\newblock Icsvr: Investigating compositional and semantic understanding in
  video retrieval models.
\newblock \emph{arXiv preprint arXiv:2306.16533}, 2023.

\bibitem[Parcalabescu et~al.(2021)Parcalabescu, Cafagna, Muradjan, Frank,
  Calixto, and Gatt]{parcalabescu2021valse}
Letitia Parcalabescu, Michele Cafagna, Lilitta Muradjan, Anette Frank, Iacer
  Calixto, and Albert Gatt.
\newblock Valse: A task-independent benchmark for vision and language models
  centered on linguistic phenomena.
\newblock \emph{arXiv preprint arXiv:2112.07566}, 2021.

\bibitem[Pham et~al.(2021)Pham, Kafle, Lin, Ding, Cohen, Tran, and
  Shrivastava]{pham2021learning}
Khoi Pham, Kushal Kafle, Zhe Lin, Zhihong Ding, Scott Cohen, Quan Tran, and
  Abhinav Shrivastava.
\newblock Learning to predict visual attributes in the wild.
\newblock In \emph{Proceedings of the IEEE/CVF conference on computer vision
  and pattern recognition}, pages 13018--13028, 2021.

\bibitem[Pratt et~al.(2020)Pratt, Yatskar, Weihs, Farhadi, and
  Kembhavi]{pratt2020grounded}
Sarah Pratt, Mark Yatskar, Luca Weihs, Ali Farhadi, and Aniruddha Kembhavi.
\newblock Grounded situation recognition.
\newblock In \emph{Computer Vision--ECCV 2020: 16th European Conference,
  Glasgow, UK, August 23--28, 2020, Proceedings, Part IV 16}, pages 314--332.
  Springer, 2020.

\bibitem[Radford et~al.(2021)Radford, Kim, Hallacy, Ramesh, Goh, Agarwal,
  Sastry, Askell, Mishkin, Clark, et~al.]{radford2021learning}
Alec Radford, Jong~Wook Kim, Chris Hallacy, Aditya Ramesh, Gabriel Goh,
  Sandhini Agarwal, Girish Sastry, Amanda Askell, Pamela Mishkin, Jack Clark,
  et~al.
\newblock Learning transferable visual models from natural language
  supervision.
\newblock In \emph{International conference on machine learning}, pages
  8748--8763. PMLR, 2021.

\bibitem[Sharma et~al.(2018)Sharma, Ding, Goodman, and
  Soricut]{sharma2018conceptual}
Piyush Sharma, Nan Ding, Sebastian Goodman, and Radu Soricut.
\newblock Conceptual captions: A cleaned, hypernymed, image alt-text dataset
  for automatic image captioning.
\newblock In \emph{Proceedings of the 56th Annual Meeting of the Association
  for Computational Linguistics (Volume 1: Long Papers)}, pages 2556--2565,
  2018.

\bibitem[Shi et~al.(2023{\natexlab{a}})Shi, Tao, Jin, Yang, Yuan, and
  Wang]{shi2023upop}
Dachuan Shi, Chaofan Tao, Ying Jin, Zhendong Yang, Chun Yuan, and Jiaqi Wang.
\newblock Upop: Unified and progressive pruning for compressing vision-language
  transformers.
\newblock In \emph{International Conference on Machine Learning}, pages
  31292--31311. PMLR, 2023{\natexlab{a}}.

\bibitem[Shi et~al.(2023{\natexlab{b}})Shi, Tao, Rao, Yang, Yuan, and
  Wang]{shi2023crossget}
Dachuan Shi, Chaofan Tao, Anyi Rao, Zhendong Yang, Chun Yuan, and Jiaqi Wang.
\newblock Crossget: Cross-guided ensemble of tokens for accelerating
  vision-language transformers.
\newblock \emph{arXiv preprint arXiv:2305.17455}, 2023{\natexlab{b}}.

\bibitem[Singh et~al.(2023)Singh, Zhang, Wang, Wang, Xiong, Du, and
  Chen]{singh2023coarse}
Harman Singh, Pengchuan Zhang, Qifan Wang, Mengjiao Wang, Wenhan Xiong, Jingfei
  Du, and Yu Chen.
\newblock Coarse-to-fine contrastive learning in image-text-graph space for
  improved vision-language compositionality.
\newblock \emph{arXiv preprint arXiv:2305.13812}, 2023.

\bibitem[Tao et~al.(2020)Tao, Jiang, Duan, and Luo]{tao2020dynamic}
Chaofan Tao, Qinhong Jiang, Lixin Duan, and Ping Luo.
\newblock Dynamic and static context-aware lstm for multi-agent motion
  prediction.
\newblock In \emph{European Conference on Computer Vision}, pages 547--563.
  Springer, 2020.

\bibitem[Thrush et~al.(2022)Thrush, Jiang, Bartolo, Singh, Williams, Kiela, and
  Ross]{thrush2022winoground}
Tristan Thrush, Ryan Jiang, Max Bartolo, Amanpreet Singh, Adina Williams, Douwe
  Kiela, and Candace Ross.
\newblock Winoground: Probing vision and language models for visio-linguistic
  compositionality.
\newblock In \emph{Proceedings of the IEEE/CVF Conference on Computer Vision
  and Pattern Recognition}, pages 5238--5248, 2022.

\bibitem[Wan et~al.(2024)Wan, Liu, Wang, Tao, Shen, Peng, Fu, Arcucci, Yao, and
  Zhang]{wan2024electrocardiogram}
Zhongwei Wan, Che Liu, Xin Wang, Chaofan Tao, Hui Shen, Zhenwu Peng, Jie Fu,
  Rossella Arcucci, Huaxiu Yao, and Mi Zhang.
\newblock Electrocardiogram instruction tuning for report generation.
\newblock \emph{arXiv preprint arXiv:2403.04945}, 2024.

\bibitem[Wang et~al.(2022)Wang, Chen, Wu, Luo, Zhou, Zhao, Xie, Liu, Jiang, and
  Yuan]{wang2022omnivl}
Junke Wang, Dongdong Chen, Zuxuan Wu, Chong Luo, Luowei Zhou, Yucheng Zhao,
  Yujia Xie, Ce Liu, Yu-Gang Jiang, and Lu Yuan.
\newblock Omnivl: One foundation model for image-language and video-language
  tasks.
\newblock \emph{Advances in neural information processing systems},
  35:\penalty0 5696--5710, 2022.

\bibitem[Wang et~al.(2023)Wang, Ge, Yan, Ge, Lin, Tsutsui, Lin, Cai, Wu, Shan,
  et~al.]{wang2023all}
Jinpeng Wang, Yixiao Ge, Rui Yan, Yuying Ge, Kevin~Qinghong Lin, Satoshi
  Tsutsui, Xudong Lin, Guanyu Cai, Jianping Wu, Ying Shan, et~al.
\newblock All in one: Exploring unified video-language pre-training.
\newblock In \emph{Proceedings of the IEEE/CVF Conference on Computer Vision
  and Pattern Recognition}, pages 6598--6608, 2023.

\bibitem[Wang and Gupta(2018)]{wang2018videos}
Xiaolong Wang and Abhinav Gupta.
\newblock Videos as space-time region graphs.
\newblock In \emph{Proceedings of the European conference on computer vision
  (ECCV)}, pages 399--417, 2018.

\bibitem[Wu et~al.(2022)Wu, Liang, Ji, Yang, Fang, Jiang, and Duan]{wu2022nuwa}
Chenfei Wu, Jian Liang, Lei Ji, Fan Yang, Yuejian Fang, Daxin Jiang, and Nan
  Duan.
\newblock N{\"u}wa: Visual synthesis pre-training for neural visual world
  creation.
\newblock In \emph{European conference on computer vision}, pages 720--736.
  Springer, 2022.

\bibitem[Xu et~al.(2019)Xu, Wong, Li, Zhao, and Kankanhalli]{xu2019learning}
Bingjie Xu, Yongkang Wong, Junnan Li, Qi Zhao, and Mohan~S Kankanhalli.
\newblock Learning to detect human-object interactions with knowledge.
\newblock In \emph{Proceedings of the IEEE/CVF Conference on Computer Vision
  and Pattern Recognition}, 2019.

\bibitem[Xu et~al.(2017)Xu, Zhu, Choy, and Fei-Fei]{xu2017scene}
Danfei Xu, Yuke Zhu, Christopher~B Choy, and Li Fei-Fei.
\newblock Scene graph generation by iterative message passing.
\newblock In \emph{Proceedings of the IEEE conference on computer vision and
  pattern recognition}, pages 5410--5419, 2017.

\bibitem[Xu et~al.(2016)Xu, Mei, Yao, and Rui]{xu2016msr}
Jun Xu, Tao Mei, Ting Yao, and Yong Rui.
\newblock Msr-vtt: A large video description dataset for bridging video and
  language.
\newblock In \emph{Proceedings of the IEEE conference on computer vision and
  pattern recognition}, pages 5288--5296, 2016.

\bibitem[Yao et~al.(2021)Yao, Huang, Hou, Lu, Niu, Xu, Liang, Li, Jiang, and
  Xu]{yao2021filip}
Lewei Yao, Runhui Huang, Lu Hou, Guansong Lu, Minzhe Niu, Hang Xu, Xiaodan
  Liang, Zhenguo Li, Xin Jiang, and Chunjing Xu.
\newblock Filip: Fine-grained interactive language-image pre-training.
\newblock \emph{arXiv preprint arXiv:2111.07783}, 2021.

\bibitem[Yu et~al.(2022)Yu, Wang, Vasudevan, Yeung, Seyedhosseini, and
  Wu]{yu2022coca}
Jiahui Yu, Zirui Wang, Vijay Vasudevan, Legg Yeung, Mojtaba Seyedhosseini, and
  Yonghui Wu.
\newblock Coca: Contrastive captioners are image-text foundation models.
\newblock \emph{arXiv preprint arXiv:2205.01917}, 2022.

\bibitem[Yuksekgonul et~al.(2022)Yuksekgonul, Bianchi, Kalluri, Jurafsky, and
  Zou]{yuksekgonul2022and}
Mert Yuksekgonul, Federico Bianchi, Pratyusha Kalluri, Dan Jurafsky, and James
  Zou.
\newblock When and why vision-language models behave like bags-of-words, and
  what to do about it?
\newblock In \emph{The Eleventh International Conference on Learning
  Representations}, 2022.

\bibitem[Zhao et~al.(2022)Zhao, Zhang, Zhu, Shen, Lee, Lu, and Yin]{zhao2022vl}
Tiancheng Zhao, Tianqi Zhang, Mingwei Zhu, Haozhan Shen, Kyusong Lee, Xiaopeng
  Lu, and Jianwei Yin.
\newblock Vl-checklist: Evaluating pre-trained vision-language models with
  objects, attributes and relations.
\newblock \emph{arXiv preprint arXiv:2207.00221}, 2022.

\end{thebibliography}
}

\clearpage
\setcounter{page}{1}
\setcounter{section}{0}
\maketitlesupplementary

%

\section{Data Specification}

\begin{table*}[htpb]
	\centering
	\scalebox{0.95}{
		\setlength{\tabcolsep}{1.2mm}{
			\begin{tabular}{l|c|cccc}
				\toprule
				\textbf{Configuration} & \textbf{Pre-training} & \multicolumn{4}{c}{\textbf{Fine-tuning}} \\
				\hline
				Optimizer &  \multicolumn{5}{c}{AdamW}  \\
				Weight Decay &  \multicolumn{5}{c}{0.2} \\
				Spatial Resolution & \multicolumn{5}{c}{224*224}   \\
				Dataset & CC3M+WebVid2M & AARO-MSRVTT & AARO-DiDeMo & AARO-ActivityNet &AARO-COCO\\
				Initial Learning Rate & 1e-4 & 1e-5 & 1e-5 & 1e-5& 1e-5\\
				Batch Size & 64 & 32 & 24 & 32 & 512\\
				GPUs & 8 & 4 & 1 & 1 &4 \\
				Epochs & 10 & 5 & 10 & 10&10 \\
				Training Frames & 4 & 12 & 12 & 12 &1\\
				Inference Frames & - & 12 & 12 & 12 &1\\
				\bottomrule
			\end{tabular}
	}}
	\caption{Configurations in the pre-training and fine-tuning stage.}
	\label{tbl:hyper}
\end{table*}

In Section \ref{sec:datasets}, we introduce the datasets used in the experiments. The detailed data specification is listed as below.  In practice,  we adopt a mixed-type negative text augmentation that use the rule-based and LLM-based augmentation with the equal possibility.

Note that if all the words in the sentence have no valid replacement in the pre-defined lists in the rule-based augmentation, we will adopt the LLM-based approach as a remedy for this sample. In addition, we view the image as the single-frame video when applying the video-text model to the image-text reasoning.

\paragraph{AARO-MSRVTT} MSRVTT \citep{xu2016msr} contains 10K video clips spanning across 41 categories, sourced primarily from real-world videos. There are totally 180k video-text pairs used for negative augmentation.
\paragraph{AARO-DiDeMo} DiDeMo \citep{anne2017localizing} focuses on the task of localizing specific moments in videos using natural language descriptions. 8.4K video-text pairs in training corpus are used for negative augmentation.
\paragraph{AARO-ActivityNet} ActivityNet \citep{heilbron2015activitynet} is a comprehensive dataset designed mainly for human activity recognition. It provides a large-scale benchmark for various tasks like activity classification, detection, and captioning. 10K video-text pairs in training corpus are used for negative augmentation.
\paragraph{VL-Checklist} VL-Checklist \citep{zhao2022vl} is a large-scale image-text compositional understanding benchmark generated from the source corpus VG \citep{krishna2017vg}, HAKE \citep{li2019hake}, SWiG \citep{pratt2020grounded} and VAW \citep{pham2021learning} which have 108K, 104K, 126K and 72K cases, respectively. The combined dataset evaluate the Attribute, Relation and Object appeared in the images. 
\paragraph{AARO-COCO} COCO caption \citep{lin2014microsoft} comprises 118K training images with 567K image-text pairs from various domains, such as urban, indoor, and outdoor settings. 118k images are divided into training set. To study whether the proposed negative augmentation can also boost image-text compositionality, we construct the dataset AARO-COCO.

\section{Implementation Details}
The training procedure is stated in the Section \ref{sec:procedure}. We adopt a two-stage video-text training, including the pre-training and fine-tuning stage. The hyper-parameters used in our experiments  are reported in the Table \ref{tbl:hyper} as suggested by the latest empirical study on the video-text pre-training \citep{cheng2023vindlu}. We build a highly performant pre-trained video-text model and then fine-tuning for the video-text and image-text reasoning tasks.

\begin{figure}[htp]
	\vspace{-1mm}
	\centering
	\includegraphics[width=0.95\linewidth]{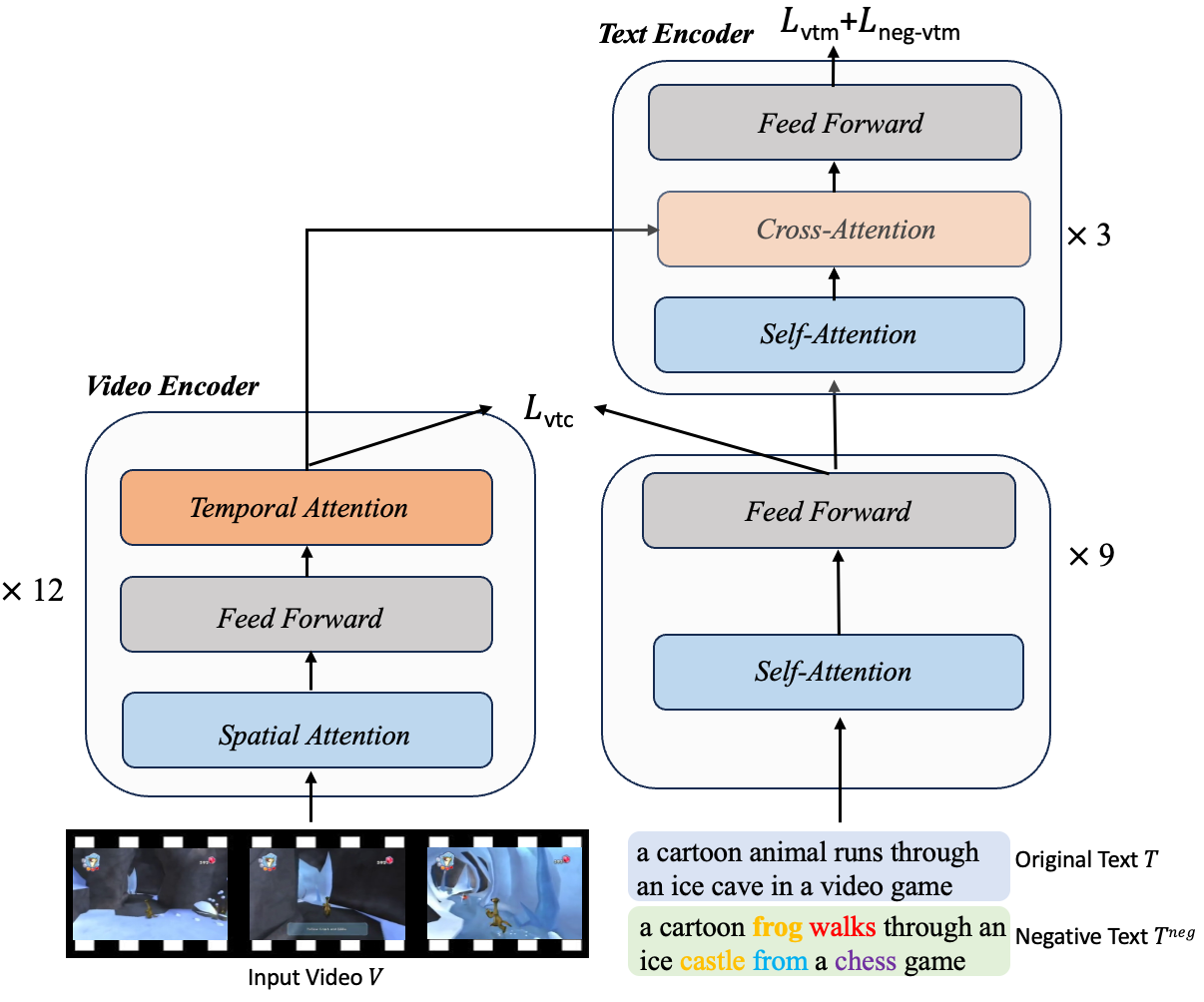}
	\caption{Illustration of the video-language model architecture.}
	\label{fig:arch}
\end{figure}

\section{Illustration of the Model Architecture}
In figure \ref{fig:arch}, we present the  architecture of the adopted video-language model. After patchifying of the video, we add the absolute temporal positional embedding as \citep{bain2021frozen} and relative spatial positional embeddings \citep{bao2021beit}. In the module of cross-attention, the output video features are projected as the key and value in the cross-attention modules of the text encoder, to enhance the interaction learning between video and text features. The architecture is suggested by the latest empirical study on the video-language learning \citep{cheng2023vindlu}. It achieves a high performance on various video-language downstream tasks, for example, text-video retrieval, video open-ended question answering and video multiple-choice question-answering.

\section{Pre-defined Word Lists for Rule-based Negative Text Augmentation}
We present the pre-fined word lists used in the rule-based negative text augmentation, which is divided into `action', `color', `size', `state', `material', `noun' and `relation' branches. Besides the word lists adopted from \citep{doveh2023teaching}, we build a  word list of relation that takes the inter-object relations into account. In addition, considering the video contains more dense action information typically compared with the image data, we construct a larger word list for action manually. The detailed words in each branch are listed in Table \ref{tbl:word-list}.

\begin{table*}[h]
	\centering
	\scalebox{0.9}{
		\begin{tabular}{@{}p{1cm}p{16cm}@{}}
			\toprule
			Action&spark, touch, empty, descend, throw, shatter, stir, highlight, load, spill, exit, vibrate, buckle, shift, crawl, cuddle, light up, fly, move, clamp, shorten, disappear, warp, crack, smolder, melt, spray, jeer, narrow, split, run, snap, clap, draw, puff up, rewind, snuggle, switch, scatter, lengthen, tip, unfold, channel, redirect, fast forward, insert, divide, translate, assemble, swipe, soak, float, transition, enter, glow, ascend, decelerate, flicker, push, stitch, click, wear, loop, kiss, strike, nestle, chip, fade out, collide, bloom, tear, unite, appear, spin, salute, sew, climb, comb, deflect, lift, merge, cross, polish, zoom out, hunt, digest, separate, invert, untie, dissolve, accelerate, blink, evade, liquefy, unload, ignite, fill, coagulate, sprout, walk, solidify, erase, resize, turn off, rotate, echo, filter, shrink, attach, shred, cover, tease, tremble, extinguish, submerge, wind down, unlock, activate, whistle, expand, dent, fold, guide, drift, signal, absorb, flip, free, underline, lower, pick up, rebound, swap, fade in, pour, wipe, bundle, hover, stretch out, increase, tap, pull, scrape, decrease, deflate, lock, contort, distort, brush, rot, gasify, quiver, ripple, bend, feed, connect, dry, hug, shrug, tie, bounce back, wobble, crystallize, zoom in, slow down, toggle, bounce, inflate, roll, block, disengage, drag, mix, punch, oscillate, sweep, drip, straighten, disassemble, jump, cheer, engage, repel, ride, swim, catch, wilt, greet, wash, consume, scrub, unravel, isolate, resume, nod, wave, trap, detach, leak, lean, fissure, flash, release, squeeze, turn on, combine, capture, uncover, wind up, widen, compress, adjust, write, slide, rock, curl up, recycle, start, dim, explode, erode, remove, interact, multiply, press down, funnel, arrive, close, farewell, scratch, grow, point, corrode, darken, bow, stop, deactivate, implode, curl, flame, dodge, drop, reflect, rust, splash, break, decay, kick, swing, fracture, depart, sink, moisten, sift, emerge, disconnect, slam, miss, type, cut, rinse, revert, replace, shake, twist, pause, illuminate, humidify, weave, avoid, dehumidify, speed up, open  \\
			\hline
			Action (old)&above, across, against, along, attached to, behind, belonging to, between, carrying, covered in, covering, eating, flying in, for, from, growing on, hanging from, has, holding, in, in front of, laying on, looking at, lying on, made of, mounted on, near, of, on, on back of, over, painted on, parked on, part of, playing, riding, says, sitting on, standing on, under, using, walking in, walking on, run, watching, wearing, wears, with\\
			\hline
			Color&teal, brown, green, black, silver, white, yellow, purple, gray, blue, orange, red, blond, concrete, cream, beige, tan, pink, maroon, olive, violet, charcoal, bronze, gold, navy, coral, burgundy, mauve, peach, rust, cyan, clay, ruby, amber\\
			\hline
			Size&long, old, extended, light, hefty, scraggy, heavy, scanty, broad, little, stout, curvy, miniature, thickset, emaciated, minute, tiny, illimitable, sizable, bulky, mammoth, strapping, enormous, obese, towering, fleshy, petite, underweight, compact, measly, teensy, grand, puny, expansive, oversize, trim, beefy, lanky, slender, gaunt, pocket-size, wee, cubby, mini, thick, full-size, pint-size, unlimited, elfin, minuscule, thin, epic, outsized, trifling, huge, scrawny, giant, portly, wide, brawny, limitless, stocky, big, large, slim, immense, skimpy, immeasurable, skeletal, colossal, meager, tall, gigantic, pudgy, extensive, overweight, tubby, cosmic, microscopic, teeny, boundless, life-size, squat, fat, paltry, undersized, bony, lean, small, chunky, massive, sturdy, great, rotund, endless, narrow, titanic, hulking, short, infinitesimal, skinny, gargantuan, plump, vast\\
			\hline
			State&tinted, dust, dirty, flip, calm, burnt, sad, dry, sunny, old, young, broken, stormy, overcast, unhappy, cloudless, adult, wet, foggy, curly, stained, rough, pale, white, fresh, good, new, first, last, long, great, little, own, other, right, big, high, different, small, large, next, early, important, few, public, bad, same, able\\
			\hline
			Material&rubber, metal, denim, wooden, cloth, silk, plastic, bamboo, stone, wicker, brick, smooth, steel, iron, silver, wool, gold, glass, helium, hydrogen, ice, lace, lead, alloy, aluminum, asbestos, ash, brass, bronze, carbon dioxide, cardboard, cement, chalk, charcoal, clay, coal, copper, cotton, dust, fiberglass, gas, leather, linen, magnesium, man-made fibers, marble, mercury, mortar, mud, nickel, nitrogen, nylon, oil, oxygen, paper, paraffin, petrol, plaster, platinum, polyester, sand, slate, smoke, soil, steam, straw, tin, uranium, water, wood, zinc\\
			\hline
			Noun&Actor, Gold, Painting, Advertisement, Grass, Parrot, Afternoon, Greece, Pencil, Airport, Guitar, Piano, Ambulance, Hair, Pillow, Animal, Hamburger, Pizza, Answer, Helicopter, Planet, Apple, Helmet, Plastic, Army, Holiday, Portugal, Australia, Honey, Potato, Balloon, Horse, Queen, Banana, Hospital, Quill, Battery, House, Rain, Beach, Hydrogen, Rainbow, Beard, Ice, Raincoat, Bed, Insect, Refrigerator, Belgium, Insurance, Restaurant, Boy, Iron, River, Branch, Island, Rocket, Breakfast, Jackal, Room, Brother, Jelly, Rose, Camera, Jewellery, Russia, Candle, Jordan, Sandwich, Car, Juice, School, Caravan, Kangaroo, Scooter, Carpet, King, Shampoo, Cartoon, Kitchen, Shoe, China, Kite, Soccer, Church, Knife, Spoon, Crayon, Lamp, Stone, Crowd, Lawyer, Sugar, Daughter, Leather, Sweden, Death, Library, Teacher, Denmark, Lighter, Telephone, Diamond, Lion, Television, Dinner, Lizard, Tent, Disease, Lock, Thailand, Doctor, London, Tomato, Dog, Lunch, Toothbrush, Dream, Machine, Traffic, Dress, Magazine, Train, Easter, Magician, Truck, Egg, Manchester, Uganda, Eggplant, Market, Umbrella, Egypt, Match, Van, Elephant, Microphone, Vase, Energy, Monkey, Vegetable, Engine, Morning, Vulture, England, Motorcycle, Wall, Evening, Nail, Whale, Eye, Napkin, Window, Family, Needle, Wire, Finland, Nest, Xylophone, Fish, Nigeria, Yacht, Flag, Night, Yak, Flower, Notebook, Zebra, Football, Ocean, Zoo, Forest, Oil, Garden, Fountain, Orange, Gas, France, Oxygen, Girl, Furniture, Oyster, Glass, Garage, Ghost\\
			\hline
			Relation &    in front of, behind, beside, above, below, next to, between, attached to, belonging to, covered in, covering, flying in, growing on, hanging from, laying on, looking at, lying on, made of, mounted on, near, on back of, over, painted on, parked on, part of, sitting on, standing on, under, walking in, walking on, wearing, wears, with, across, against, along, carrying, for, from, has, holding, in, on, of, playing, riding, says, using, watching, hover over\\
			
			\hline
		\end{tabular}
	}
	\caption{Word lists for the rule-based negative text augmentation}
	\label{tbl:word-list}
\end{table*}

\end{document}